\documentclass[11pt,letterpaper]{article}

\usepackage{cvpr}
\usepackage{times}
\usepackage{epsfig}
\usepackage{graphicx}
\usepackage{amsmath}
\usepackage{amssymb}
\usepackage{authblk}

\usepackage[breaklinks=true,bookmarks=false]{hyperref}

\cvprfinalcopy 

\def\cvprPaperID{****} 

\begin{document}

\title{The Rapidly Changing Landscape of Conversational Agents}


\author{\textbf{Vinayak Mathur}}
\author{\textbf{Arpit Singh}}
\affil{College of Information and Computer Sciences\\University of Massachusetts Amherst \\ \tt\small \{vinayak, arpitsingh\}@cs.umass.edu}
\maketitle

\begin{abstract}
Conversational agents have become ubiquitous, ranging from goal-oriented systems for helping with reservations to chit-chat models found in modern virtual assistants. In this survey paper, we explore this fascinating field. We look at some of the pioneering work that defined the field and gradually move to the current state-of-the-art models. We look at statistical, neural, generative adversarial network based and reinforcement learning based approaches and how they evolved. Along the way we discuss various challenges that the field faces, lack of context in utterances, not having a good quantitative metric to compare models, lack of trust in agents because they do not have a consistent persona etc. We structure this paper in a way that answers these pertinent questions and discusses competing approaches to solve them.
\end{abstract}

\section{Introduction}
One of the earliest goals of Artificial Intelligence (AI) has been to build machines that can converse with us. Whether in early AI literature or the current popular culture, conversational agents have captured our imagination like no other technology has. In-fact the ultimate test of whether true artificial intelligence has been achieved, the Turing test \cite{turing} proposed by Alan Turing the father of artificial intelligence in 1950, revolves around the concept of a good conversational agent. The test is deemed to have been passed if a conversational agent is able to fool human judges into believing that it is in fact a human being. 
\\ \\
Starting with pattern matching programs like ELIZA developed at MIT in 1964 to the current commercial conversational agents and personal assistants (Siri, Allo, Alexa, Cortana et al) that all of us carry in our pockets, conversational agents have come a long way. In this paper we look at this incredible journey. We start by looking at early rule-based methods which consisted of hand engineered features, most of which were domain specific. However, in our view, the advent of neural networks that were capable of capturing long term dependencies in text and the creation of the sequence to sequence learning model \cite{seq} that was capable of handling utterances of varying length is what truly revolutionized the field. Since the sequence to sequence model was first used to build a neural conversational agent \cite{vinyals2015neural} in 2016 the field has exploded. With a multitude of new approaches being proposed in the last two years which significantly impact the quality of these conversational agents, we skew our paper towards the post 2016 era. Indeed one of the key features of this paper is that it surveys the exciting new developments in the domain of conversational agents. \\ \\
Dialogue systems, also known as interactive conversational agents, virtual agents and sometimes chatterbots, are used in a wide set of applications ranging from technical support services to language learning tools and entertainment. Dialogue systems can be divided into goal-driven systems, such as technical support services, booking systems, and querying systems. On the other hand we have non-goal-driven systems which are also referred to as chit-chat models. There is no explicit purpose for interacting with these agents other than entertainment. Compared to goal oriented dialog systems where the universe is limited to an application, building open-ended chit-chat models is more challenging. Non-goal oriented agents are a good indication of the state of the art of artificial intelligence according to the Turing test.  With no grounding in common sense and no sense of context these agents have to fall back on canned responses and resort to internet searches now. But as we discuss in section \ref{kb} , new techniques are emerging to provide this much needed context to these agents. \\ \\
The recent successes in the domain of Reinforcement Learning (RL) has also opened new avenues of applications in the conversational agent setting. We explore some of these approaches in section \ref{rl}\\ \\
Another feature that has been traditionally lacking in conversation agents is a personality. O Vinayal et al \cite{vinyals2015neural} hypothesis that not having a consistent personality is one of the main reasons that is stopping us from passing the turing test. Conversational agents also lack emotional consistency in their responses. These features are vital if we want humans  to trust conversational agents. In section \ref{human} we discuss state of the art approaches to overcome these problems.\\ \\
Despite such huge advancements in the field, the way these models are evaluated is something that needs to be dramatically altered. Currently there exists no perfect quantitative method to compare two conversational agents. The field has to rely on qualitative measures or measures like BLeU and perplexity borrowed from machine translation. In section \ref{eval} we discuss this problem in detail.

\section{Early Techniques}

\begin{figure}[h]
    \centering
    \includegraphics{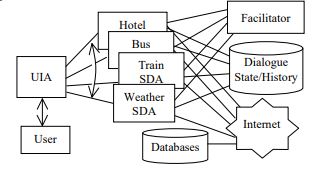}
    \caption{Agent society for spoken dialogue for tour information service \cite{distarch}}
    \label{fig:mesh1}
\end{figure}

Initially, the interactive dialogue systems were based on and limited to speaker independent recognition of isolated words and phrases or limited continuous speech such as digit strings. In August 1993, there came the ESPRIT SUNDIAL project (Peckham et al, 1993 \cite{sundial}) which was aimed at allowing spontaneous conversational inquiries over the telephone for the train timetable and flight enquiries.  The linguistic processing component in it was based on natural language parsing. The parser made use of alternative word hypotheses represented in a lattice or graph in constructing a parse tree and allowance was made for gaps and partially parsable strings. It made use of both syntactic and semantic knowledge for the task domain. It was able to achieve a 96\% success rate for the flight inquiry application in English. However, the issue was that the given conversational agent was heavily limited to the types of applications it can perform and its high success rate was more due to that instead of great natural language techniques (relative to recent times). 
\\ \\
In 1995, two researchers (Ball et al, 1995 \cite{persona}) at Microsoft developed a conversational assistant called Persona which was one of the first true personal assistant similar to what we have in recent times (like Siri, etc). It allowed users the maximum flexibility to express their requests in whatever syntax they found most natural and the interface was based on a broad-coverage NLP system unlike the system discussed in the previous paragraph. In this, a labelled semantic graph is generated from the speech input which encodes case frames or thematic roles. After this, a sequence of graph transformations is applied on it using the knowledge of interaction scenario and application domain. This results into a normalized application specific structure called as task graph which is then matched against the templates (in the application) which represent the normalized task graphs corresponding to all the possible user statements that the assistant understands and the action is then executed. The accuracy was not that good and they did not bother to calculate it. Also, due to the integrated nature of conversational interaction in Persona, the necessary knowledge must be provided to each component of the system. Although it had limitations, it provided a very usable linguistic foundation for conversational interaction. 
\\ \\
The researchers thought that if they can create assistant models specific to the corresponding models, they can achieve better accuracy for those applications instead of creating a common unified personal assistant which at that time performed quite poorly. There was a surge in application-specific assistants like in-car intelligent personal assistant (Schillo et al, 1996 \cite{incar}), spoken-language interface to execute military exercises (Stent et al, 1999 \cite{commandtalk}), etc. Since it was difficult to develop systems with high domain extensibility, the researchers came up with a distributed architecture for cooperative spoken dialogue agents (Lin et al, 1999 \cite{distarch}). 
\\ \\
Under this architecture, different spoken dialogue agents handling different domains can be developed independently and cooperate with one another to respond to the user’s requests. While a user interface agent can access the correct spoken dialogue agent through a domain switching protocol, and carry over the dialogue state and history so as to keep the knowledge processed persistently and consistently across different domains. Figure \ref{fig:mesh1} shows the agent society for spoken dialogue for tour information service.
\\ \\

If we define the false alarm rate by counting the utterances in which unnecessary domain-switching occurred and the detection rate by counting the utterances in which the desired domain-switching were accurately detected, then in this model, high detection rate was achieved at very low false alarm rate. For instance, for around a false alarm rate of 0.2, the model was able to achieve a detection rate of around 0.9 for the case of tag sequence search with language model search scheme.

\section{Machine Learning Methods}

Next came the era of using machine learning methods in the area of conversation agents which totally revolutionized this field. 
\\ \\
Maxine Eskenazi and her team initially wanted to build spoken dialog system for the less general sections of the population, such as the elderly and non-native speakers of English. They came up with Let’s Go project (Raux et al, 2003 \cite{alan1}) that was designed to provide Pittsburgh area bus information. Later, this was opened to the general public (Raux et al, 2005 \cite{alan2}). Their work is important in terms of the techniques they used.
\\ \\
The speech recognition was done using n-gram statistical model which is then passed to a robust parser based on an extended Context Free Grammar allowing the system to skip unknown words and perform partial parsing. They wrote the grammar based on a combination of their own intuition and a small scale Wizard-of-Oz experiment they ran. The grammar rules used to identify bus stops were generated automatically from the schedule database. After this, they trained a statistical language model on the artificial corpus. In order to make the parsing grammar robust enough to parse fairly ungrammatical, yet understandable sentences, it was kept as general as possible. On making it public, they initially achieved a task success rate of 43.3\% for the whole corpus and 43.6\ when excluding sessions that did not contain any system-directed speech.
\\ \\
After this they tried to increase the performance of the system (Raux et al, 2006 \cite{alan3}). They retrained their acoustic models by performing Baum-Welch optimization on the transcribed data (starting from their original models). Unfortunately, this only brought marginal improvement because the models (semi-continuous HMMs) and algorithms they were using were too simplistic for this task. They improved the turn-taking management abilities of the system by closely analysing the feedback they received. They added more specific strategies, aiming at dealing with problems like noisy environments, too loud or too long utterances, etc. They found that they were able to get a success rate of 79\% for the complete dialogues (which was great). 
\\ \\
The previous papers (like the ones which we discussed in the above paragraph) did not attempt to use data-driven techniques for the dialog agents because such data was not available in large amount at that time. But then there was a high increase in the collection of spoken dialog corpora which made it possible to use data-driven techniques to build and use models of task-oriented dialogs and possibly get good results. In the paper by Srinivas et al,2008 \cite{mlbangalore}, the authors proposed using data-driven techniques to build task structures for individual dialogs and use the dialog task structures for dialog act classification, task/subtask classification, task/subtask prediction and dialog act prediction. 
\\ \\
For each utterance, they calculated features like n-grams of the words and their POS tags, dialog act and task/subtask label. Then they put those features in the binary MaxEnt classifier. For this, their model was able to achieve an error rate of 25.1\% for the dialog act classification which was better than the best performing models at that time. Although, according to the modern standards, the results are not that great but the approach they suggested (of using data to build machine learning models) forms the basis of the techniques that are currently used in this area.

\section{Neural Models}
\subsection{Sequence to Sequence approaches for dialogue modelling}
\begin{figure}[h]
    \centering
    \includegraphics[width=12cm, height=4cm]{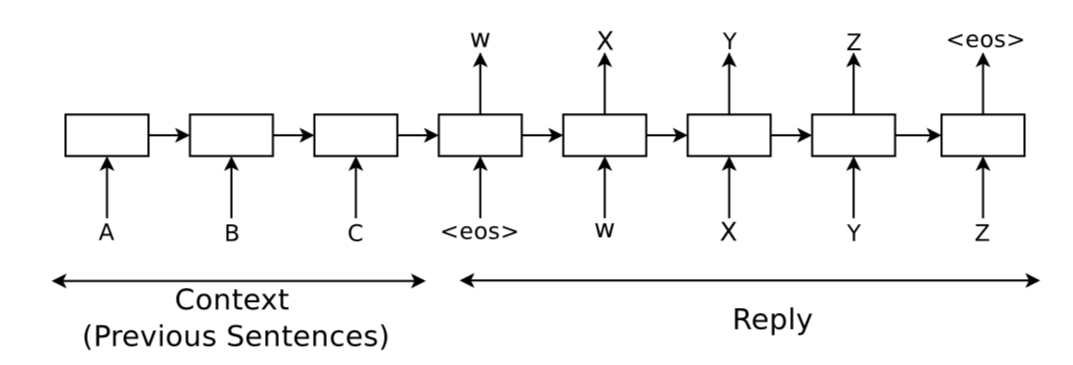}
    
    \caption{sequence to sequence framework for modelling conversation \cite{vinyals2015neural}}
    \label{fig:seq}
\end{figure}

The problem with rule-based models was that they were often domain dependent and could not be easily ported to a new domain. They also depended on hand crafted rules which was both expensive and required domain expertise. Two factors which when combined spell doom for scalbility. All of this changed in 2015 when Vinyals et al proposed an approach \cite{vinyals2015neural} inspired from the recent progress in machine translation \cite{seq}. Vinyals et al used the sequence to sequence learning architecture for conversation agents. Their model was the first model which could be trained end-to-end, and could generate a new output utterance based on just the input sentence and no other hand crafted features. 
\\
\\
They achieved this by casting the conversation modelling task, as a task of predicting the next sequence given the previous sequence using recurrent networks. This simple approach truly changed the conversation agent landscape. Most of the state-of-the-art today is built on their success. In a nutshell the input utterance is input to an encoder network, which is a recurrent neural network (RNN) in this case, but as we will see Long Short Term Memory (LSTMs) \cite{lstm} have since replaced RNNs as the standard for this task. The encoder summarizes the input utterance into a fixed length vector representation which is input to the decoder, which itself is again a RNN. The paper looks at this fixed vector as the \textit{thought vector} - which hold the most important information of the input utterance. The Decoder netwroks takes this as input and output's an output utterance word-by-word until it generates an end-of-speech  $<eos>$ token. This approach allows for variable length inputs and outputs. The network is jointly trained on two turn conversations. Figure \ref{fig:seq} shows the sequence to sequence neural conversation model. 
\\
\\
Even though most of the modern work in the field is built on this approach there is a significant drawback to this idea. This model can theoretically never \textit{solve} the problem of modelling dialogues due to various simplifications, the most important of them being the objective function that is being optimized does not capture the actual objective achieved through human communication, which is typically longer term and based on exchange of information rather than next step prediction. It is important to see that optimizing an agent to generate text based on what it sees in the two-turn conversation dataset that it is trained on does not mean that the agent would be able to generalize to human level conversation across contexts. Nevertheless in absence of a better way to capture human communication this approach laid the foundation of most of the modern advances in the field. Another problem that plagues this paper and the field in general is Evaluation. As there can be multiple correct output utterances for a given input utterance there is no quantitative way to evaluate how well a model is performing. In this paper to show the efficacy of their model the authors publish snippets of conversations across different datasets. We discuss this general problem in evaluation later. \\

\begin{figure}[h]
    \centering
    \includegraphics[width=14cm, height=7cm]{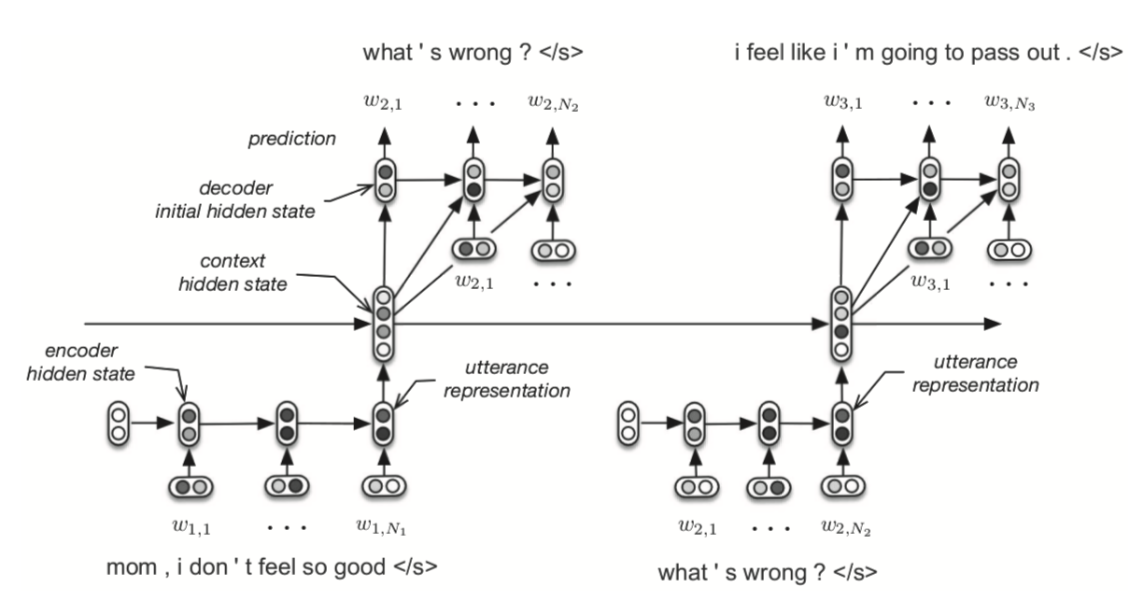}
    
    \caption{Hierarchical approach to dialogue modelling. A context RNN summarizes the utterances until that point from the encoder. The decoder produces output utterances based on the hidden state of the context RNN instead of the encoder RNN \cite{serban2016building}}
    \label{fig:hred}
\end{figure}

Iulian et al. build on this sequence-to-sequence based approach in their paper presented in AAAI 2016 \cite{serban2016building}. Their work is inspired by  the hierarchical recurrent encoder-decoder architecture (HRED) proposed by Sordoni et al. \cite{sordoni2015hierarchical}. Their premise is that a dialogue can be seen as a sequence of utterances which, in turn, are sequences of tokens. Taking advantage of this built in hierarchy they model their system in the following fashion. \\ \\
The encoder RNN maps each utterance to an utterance vector. The utterance vector is the hidden state obtained after the last token of the utterance has been processed. The higher-level context RNN keeps track of past utterances by processing iteratively each utterance vector. After processing utterance $U_m$, the hidden state of the context RNN represents a summary of the dialogue up to and including turn $m$, which is used to predict the next utterance $U_{m+1}$. The next utterance prediction is performed by means of a decoder RNN, which takes the hidden state of the context RNN and produces a probability distribution over the tokens in the next utterance. As seen in figure \ref{fig:hred}

The advantages of using a hierarchical representation are two-fold. First, the context RNN allows the model to represent a form of common ground between speakers, e.g. to represent topics and concepts shared between the speakers using a distributed vector representation. Second, because the number of computational steps between utterances is reduced. This makes the objective function more stable w.r.t. the model parameters, and helps propagate the training signal for first-order optimization methods.
\\ \\
Models like sequence-to-sequence and the hierarchical approaches have proven to be good baseline models. In the last couple of years there has been a major effort to build on top of these baselines to make conversational agents more robust \cite{r1} \cite{r2}.\\
Due to their large parameter space, the estimation of neural conversation models requires considerable amounts of dialogue data. Large online corpora are helpful for this. However  several dialogue corpora, most notably those extracted from subtitles, do not include any explicit turn segmentation or speaker identification.The neural conversation model may therefore inadvertently learn responses that remain within the same dialogue turn instead of starting a new turn. Lison et al \cite{lison2017not} overcome these limitations by introduce a weighting model into the neural architecture. The weighting model, which is itself estimated from dialogue data, associates each training example to a numerical weight that reflects its intrinsic quality for dialogue modelling. At training time, these sample weights are included into the empirical loss to be minimized. The purpose of this model is to associate each ⟨context, response⟩ example pair to a numerical weight that reflects the intrinsic “quality” of each example. The instance weights are then included in the empirical loss to minimize when learning the parameters of the neural conversation model. The weights are themselves computed via a neural model learned from dialogue data. Approaches like \cite{lison2017not} are helpful but data to train these neural conversational agents remains scarce especially in academia, we talk more about the scarcity of data in a future section.

\subsection{Language Model based approaches for dialogue modelling}
Though sequence-to-sequence based models have achieved a lot of success, another push in the field has been to instead train a
language model over the entire dialogue as one single sequence \cite{luan2016lstm}. These works argue that a language model is better suited to dialogue modeling, as it learns how the conversation evolves as information progresses. \\ \\
Mei et al. \cite{mei2017coherent} improve the coherence of such neural dialogue language models by developing a generative dynamic attention mechanism that allows each generated word to choose which related words it wants to align to in the increasing conversation history (including the previous words in the response being generated). They introduce a dynamic attention mechanism to a RNN language model in which the scope of attention increases as the recurrence operation progresses from the start through the end of the conversation. The dynamic attention model promotes coherence of the generated dialogue responses (continuations) by favoring the generation of words that have syntactic or semantic associations with salient words in the conversation history.

\section{Knowledge augmented models}
\label{kb}
Although these neural models are really powerful, so much so that they power most of the commercially available smart assistants and conversational agents. However these agents lack a sense of context and a grounding in common sense that their human interlocutors 
possess. This is especially evident when interacting with a commercial conversation agent, when more often that not the agent has to fall back to canned responses or resort to displaying Internet search results in response to an input utterance. One of the main goals of the research community, over the last year or so, has been to overcome this fundamental problem with conversation agents. A lot of different approaches have been proposed ranging from using knowledge graphs \cite{kg} to augment the agent's knowledge to using latest advancements in the field of online learning \cite{cont}. In this section we discuss some of these approaches. \\

\begin{figure}[h]
    \centering
    \includegraphics[width=14cm, height=6cm]{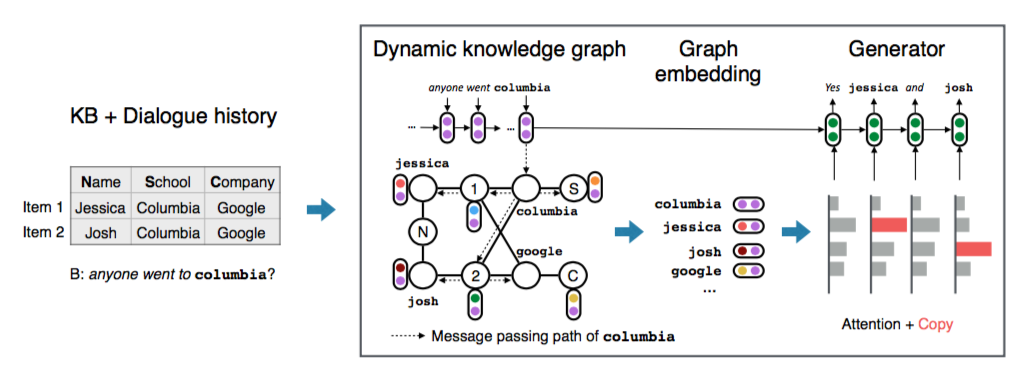}
    \caption{Example demonstrating how DynoNet augments the conversation \cite{kg}}
    \label{fig:dyno}
\end{figure}

The first approach we discuss is the Dynamic Knowledge Graph Network (DynoNet) proposed by He et al \cite{kg}, in which the dialogue state is modeled as a knowledge graph with an embedding for each node. To model both structured and open-ended context they model two agents, each with a private list of items with attributes, that must communicate to identify the unique shared item. They structure entities as a knowledge graph; as the dialogue proceeds, new nodes are added and new context is propagated on the graph. An attention-based mechanism over the node embeddings drives generation of new utterances. The model is best explained by the example used in the paper which is as follows: The knowledge graph represents entities and relations in the agent’s private KB, e.g., \textit{item-1’s} company is \textit{google}. As the conversation unfolds, utterances are embedded and incorporated into node embeddings of mentioned entities. For instance, in Figure \ref{fig:dyno}, \textit{“anyone went to columbia”} updates the embedding of \textit{columbia}. Next, each node recursively passes its embedding to neighboring nodes so that related entities (e.g., those in the same row or column) also receive information from the most recent utterance. In this example, \textit{jessica} and \textit{josh} both receive new context when \textit{columbia} is mentioned. Finally, the utterance generator, an LSTM, produces the next utterance by attending to the node embeddings.\\

\begin{figure}[h]
    \centering
    \includegraphics[width=10cm, height=5cm]{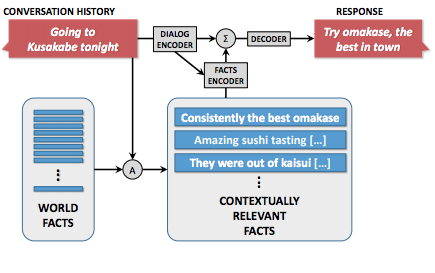}
    
    \caption{The neural architecture of the knowledge grounded model which uses a set of external world facts to augment the output utterance generated bt the model \cite{ghazvininejad2017knowledge}}
    \label{fig:kg}
\end{figure}

However Lee et al in \cite{cont} take a different approach to add knowledge to conversational agents. They proposes using a continuous learning based approach. They introduce a task-independent conversation model and an adaptive online algorithm for continual learning which together allow them to sequentially train a conversation model over multiple tasks without forgetting earlier tasks.
\\ \\
In a different approach, Ghazvininejad et al \cite{ghazvininejad2017knowledge} propose a knowledge grounded approach which infuses the output utterance with factual information relevant to the conversational context. Their architecture is shown in figure \ref{fig:kg}. They use an external collection of world facts which is a large collection of raw text entries (e.g., Foursquare, Wikipedia, or Amazon reviews) indexed by named entities as keys. Then, given a conversational history or source sequence S, they identify the “focus” in S, which is the text span (one or more entities) based on which they form a query to link to the facts. The query is then used to retrieve all contextually relevant facts. Finally, both conversation history and relevant facts are fed into a neural architecture that features distinct encoders for conversation history and facts. Another interesting facet of such a model is that new facts can be added and old facts updated by just updating the world facts dictionary without retraining the model from scratch, thus making the model more adaptive and robust.

Instead of just having a set of facts to augment the conversation, a richer way could be to use knowledge graphs or commonsense  knowledge bases which consist of [entity-relation-entity] triples. Young et al explore this idea in \cite{young2017augmenting}.
For a given input utterance, they find the relevant assertions in the common sense knowledge base using simple n-gram matching. They then perform chunking on the relevant assertions and feed the individual token to a tri-LSTM encoder. The output of this encoder is weighted along with the input utterance and the output utterance is generated. They claim that such common sense conversation agents outperform a naive conversation agent.
\\ \\
Another interesting way to add knowledge to the conversation agents is to capture external knowledge for a given dialog using a search engine. In the paper by Long et al, 2017 \cite{longsearch}, the authors built a model to generate natural and informative responses for customer service oriented dialog incorporating external knowledge. 
\\ \\
They get the external knowledge using a search engine. Then a knowledge enhanced sequence-to-sequence framework is designed to model multi-turn dialogs on external knowledge conditionally. For this purpose, their model extends the simple sequence-to-sequence model by augmenting the input with the knowledge vector so as to take account of the knowledge in the procedure of response generation into the decoder of the sequence-to-sequence model. Both the encoder and the decoder are composed of LSTM.
\\ \\
Their model scores an average human rating of 3.3919 out of 5 in comparison to the baseline which is 3.3638 out of 5. Hence, their model generates more informative responses. However, they found the external knowledge plays a negative role in the procedure of response generation when there is more noise in the information. Exploring how to obtain credible knowledge of a given dialog history can be a future generation of their model.

\section{Reinforcement Learning based models}
\label{rl}

After exploring the neural methods in a lot of detail, the researchers have also begun exploring, in the current decade, how to use the reinforcement learning methods in the dialogue and personal agents. 

\subsection{Initial reinforcement methods}

One of the first main papers that thought of using reinforcement learning for this came in 2005 by English et al \cite{rlintro}. They used an on-policy Monte Carlo method and the objective function they used was a linear combination of the solution quality (S) and the dialog length (L), taking the form: o(S,I) = $w_1S$ - $w_2L $.
\\ \\
At the end of each dialog the interaction was given a score based on the evaluation function and that score was used to update the dialog policy of both agents (that is, the conversants). The state-action history for each agent was iterated over separately and the score from the recent dialog was averaged in with the expected return from the existing policy. They chose not to include any discounting factor to the dialog score as they progressed back through the dialog history. The decision to equally weight each state-action pair in the dialog history was made because an action’s contribution to the dialog score is not dependent upon its proximity to the end of the task. In order to combat the problem of converging to an effective policy they divided up the agent training process into multiple epochs. 
\\ \\
The average objective function score for the case of learned policies was 44.90. One of the main reasons for the low accuracy (which is also a limitation of this paper) was that there were a number of aspects of dialog that they had not modeled such as non-understandings, misunderstandings, and even parsing sentences into the action specification and generating sentences from the action specification. But the paper set the pavement of the reinforcement learning methods into the area of dialog and personal agents.

\subsection{End-to-End Reinforcement Learning of Dialogue Agents for Information Access}

Let’s have a look at KB-InfoBot (by Dhingra et al, 2017 \cite{rlendtoend}): a multi-turn dialogue agent which helps users search Knowledge Bases (KBs) without composing complicated queries. In this paper, they replace the symbolic queries (which break the differentiability of the system and prevent end-to-end training of neural dialogue agents) with an induced ‘soft’ posterior distribution over the KB that indicates which entities the user is interested in. Integrating the soft retrieval process with a reinforcement learner leads to higher task success rate and reward in both simulations and against real users.
\\ \\
In this, the authors used an RNN to allow the network to maintain an internal state of dialogue history. Specifically, they used a Gated Recurrent Unit followed by a fully-connected layer and softmax non-linearity to model the policy π over the actions. During training, the agent samples its actions from this policy to encourage exploration. Parameters of the neural components were trained using the REINFORCE algorithm. For end-to-end training they updated both the dialogue policy and the belief trackers using the reinforcement signal. While testing, the dialogue is regarded as a success if the user target is in top five results returned by the agent and the reward is accordingly calculated that helps the agent take the next action.
\\ \\
Their system returns a success rate of 0.66 for small knowledge bases and a great success rate of 0.83 for medium and large knowledge bases. As the user interacts with the agent, the collected data can be used to train the end-to-end agent which we see has a strong learning capability. Gradually, as more experience is collected, the system can switch from Reinforcement Learning-Soft to the personalized end-to-end agent. Effective implementation of this requires such personalized end-to-end agents to learn quickly which should be explored in the future. 
\\ \\
However, the system has a few limitations. The accuracy is not enough for using for the practical applications. The agent suffers from the cold start issue. In the case of end-to-end learning, they found that for a moderately sized knowledge base, the agent almost always fails if starting from random initialization.

\subsection{Actor-Critic Algorithm}

Deep reinforcement learning (RL) methods have significant potential for dialogue policy optimisation. However, they suffer from a poor performance in the early stages of learning as we saw in the paper in the above section. This is especially problematic for on-line learning with real users.
\\ \\
In the paper by Su et al, 2017 \cite{rlactorcritic}, they proposed a sample-efficient actor-critic reinforcement learning with supervised data for dialogue management. Just for a heads up, actor-critic algorithms are the algorithms that have an actor stores the policy according to which the action is taken by the agent and a critic that critiques the actions chosen by the actor (that is, the rewards obtained after the action are sent to the critic using which it calculates value functions).
\\ \\
To speed up the learning process, they presented two sample-efficient neural networks algorithms: trust region actor-critic with experience replay (TRACER) and episodic natural actor-critic with experience replay (eNACER). Both models employ off-policy learning with experience replay to improve sample-efficiency.  For TRACER, the trust region helps to control the learning step size and avoid catastrophic model changes. For eNACER, the natural gradient identifies the steepest ascent direction in policy space to speed up the convergence. 
\\ \\
To mitigate the cold start issue, a corpus of demonstration data was utilised to pre-train the models prior to on-line reinforcement learning. Combining these two approaches, they demonstrated a practical approach to learn deep RL-based dialogue policies and also demonstrated their effectiveness in a task-oriented information seeking domain.

\begin{figure}[h]
    \centering
    \includegraphics{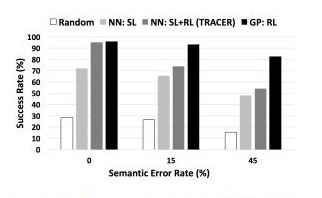}
    \caption{The success rate of TRACER for a random policy, policy trained with corpus data (NN:SL) and further improved via RL (NN:SL+RL) respectively in user simulation under various semantic error rates \cite{distarch}}
    \label{fig:mesh7}
\end{figure}

We can see in the figure \ref{fig:mesh7} that the success rate reaches at around 95\% for the case of policy trained with corpus data and using reinforcement learning which is impressive. Also, they train very quickly. For instance, for training just around 500-1000 dialogues, eNACER has a success rate of around 95\% and TRACER has a success rate of around 92\%. However, the authors noted that performance falls off rather rapidly in noise as the uncertainty estimates are not handled well by neural networks architectures. This can also be a topic for future research.

\subsection{Using Generative Adversarial Network}

Recently, generative adversarial networks are being explored and how they can be used in the dialog agents. Although generative adversarial networks are a topic in itself to explore. However, the paper mentioned below used uses reinforcement learning along with generative adversarial network so we cover it here inside the reinforcement learning methods. They can be used by the applications to generate dialogues similar to humans.
\\ \\
In the paper by Li et al, 2017 \cite{rlgan}, the authors proposed using adversarial training for open-domain dialogue generation such that the system is trained to produce sequences that are indistinguishable from human-generated dialogue utterances. The task is considered as a reinforcement learning problem where two systems get jointly trained: a generative model to produce response sequences, and a discriminator (similar to the human evaluator in the Turing test) that distinguishes between the human-generated dialogues and the machine-generated ones. The generative model defines the policy that generates a response given the dialog history and the discriminative model is a binary classifier that takes a sequence of dialog utterances as inputs and outputs whether the input is generated by the humans or machines. The outputs from the discriminator are then used as rewards for the generative model pushing the system to generate dialogues that mostly resemble human dialogues. 
\\ \\
The key idea of the system is to encourage the generator to generate utterances that are indistinguishable from human generated dialogues. The policy gradient methods are used to achieve such a goal, in which the score of current utterances being human-generated ones assigned by the discriminator is used as a reward for the generator, which is trained to maximize the expected reward of generated utterances using the REINFORCE algorithm.
\\ \\
Their model achieved a machine vs random accuracy score of 0.952 out of 1. However, on applying the same training paradigm to machine translation in preliminary experiments, the authors did not find a clear performance boost. They thought that it may be because the adversarial training strategy is more beneficial to tasks in which there is a big discrepancy between the distributions of the generated sequences and the reference target sequences (that is, the adversarial approach may be more beneficial on tasks in which entropy of the targets is high). In the future, this relationship can be further explored.

\section{Approaches to Human-ize agents}
\label{human}
A lack of a coherent personality in conversational agents that most of these models propose has been identified as one of the primary reasons that these agents have not been able to pass the Turing test \cite{turing} \cite{vinyals2015neural}. Aside from such academic motivations, making conversational agents more like their human interlocutors which posses both a persona and are capable of parsing emotions is of great practical and commercial use. Consequently in the last couple of years different approaches have been tried to achieve this goal.\\ \\
Li et al \cite{li2016persona} address the challenge of consistency and how to endow data-driven systems with the coherent “persona” needed to model human-like behavior. They consider a  persona to be composite of elements of identity (background facts or user
profile), language behavior, and interaction style. They also account for a persona to be adaptive since an agent may
need to present different facets to different human interlocutors depending on the interaction. Ultimately these personas are incorporated into the model as embeddings. 
Adding a persona not only improves the human interaction but also improves BLeU score and perplexity over the baseline sequence to sequence models. 
The model represents each individual speaker as a vector or embedding, which encodes speaker-specific information (e.g.dialect, register, age, gender, personal information) that influences the content and style of her responses. Most importantly these traits do not need to be explicitly annotated, which would be really tedious and limit the applications of the model. Instead the model manages
to cluster users along some of these traits (e.g. age, country of residence) based on the responses alone. The model first encodes message $S$ into a vector representation $h_S$ using the source LSTM. Then for each step in the target side, hidden
units are obtained by combining the representation produced by the target LSTM at the previous time step, the word representations at the current time step, and the speaker embedding $v_i$. In this way, speaker information is encoded and injected into the hidden layer at each time step and thus helps predict personalized responses throughout the generation process. The process described here is visualizes in figure \ref{fig:per} below.\\

\begin{figure}[h]
    \centering
    \includegraphics[width=14cm, height=6cm]{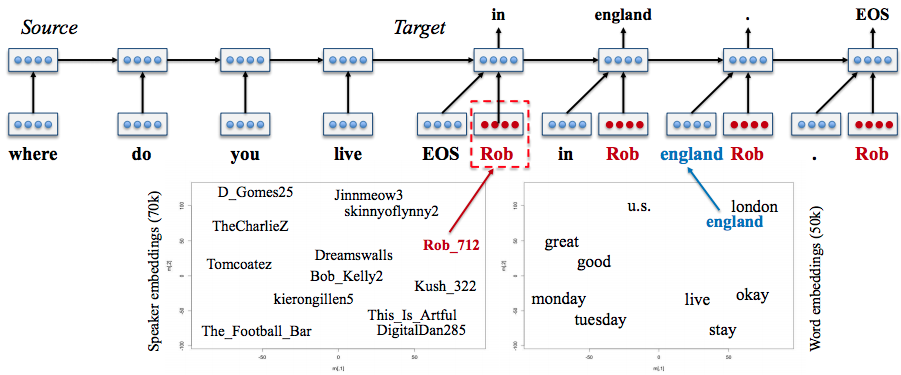}
    
    \caption{Visualization of how the persona is integrated in a sequence to sequence style conversational agent \cite{li2016persona}}
    \label{fig:per}
\end{figure}

Building on works like this the Emotional Chatting Machine model proposed by Zhou et al \cite{zhou2017emotional} is a model which generates responses that are not only grammatically consistent but also emotionally consistent. To achieve this their approach models the high-level abstraction of emotion expressions by embedding emotion categories. They also capture the change of implicit internal emotion states and use explicit emotion expressions with an external emotion vocabulary.\\ \\
Although they did not evaluate their model on some standard metric, they showed that their model can generate responses appropriate not only in content but also in emotion. In the future, instead of specifying an emotion class, the model should decide the most appropriate emotion category for the response. However, this may be challenging since such a task depends on the topic, context or the mood of the user. \\ \\
The goal of capturing emotions and having consistent personalities for a conversational agent is an important one. The field is still nascent but advances in the domain will have far reaching consequences for conversational models in general. People tend to trust agents that are emotionally consistent, and in the long term trust is what will decide the fate of large scale adoption of conversational agents.
\section{Evaluation methods}
\label{eval}
Evaluating conversational agents is an open research problem in the field. With the inclusion of emotion component in the modern conversation agents, evaluating such models has become even more complex.The current evaluation methods like perplexity and BLEU score are not good enough and correlate very weakly with human
judgments. In the paper by Liu et al, 2016 \cite{eval}, the authors discuss about how not to evaluate the dialogue system. They provide quantitative and qualitative results highlighting specific weaknesses in existing metrics and provide recommendations for the future development of better automatic evaluation metrics for dialogue systems.
\\ \\
According to them, the metrics (like Kiros et al, 2015 \cite{kiros}) that are based on distributed sentence representations hold the most promise for the future. It is because word-overlap metrics like BLEU simply require too many ground-truth responses to find a significant match for a reasonable response due to the high diversity of dialogue responses. Similarly, the metrics that are embedding-based consist of basic averages of vectors obtained through distributional semantics and so they are also insufficiently complex for modeling sentence-level compositionality in dialogue. 
\\ \\
The metrics that take into account the context can also be considered. Such metrics can come in the form of an evaluation model that is learned
from data. This model can be either a discriminative model that attempts to distinguish between model and human responses or a model that uses data collected from the human survey in order to provide human-like scores to proposed responses.
\section{Conclusion}
In this survey paper we explored the exciting and rapidly changing field of conversational agents. We talked about the early rule-based methods that depended on hand-engineered features. These methods laid the ground work for the current models. However these models were expensive to create and the features depended on the domain that the conversational agent was created for. It was hard to modify these models for a new domain. As computation power increased, and we developed neural networks that were able to capture long range dependencies (RNNs,GRUs,LSTMs) the field moved towards neural models for building these agents. Sequence to sequence model created in 2015 was capable of handling utterances of variable lengths, the application of sequence to sequence to conversation agents truly revolutionized the domain. After this advancement the field has literally exploded with numerous application in the last couple of years. The results have been impressive enough to find their way into commercial applications such that these agents have become truly ubiquitous. We attempt to present a broad view of these advancements with a focus on the main challenges encountered by the conversational agents and how these new approaches are trying to mitigate them.

{\small
\bibliographystyle{ieeetr}
\bibliography{egpaper_final}
}

\end{document}